| PAPER TITLE | INTELLIGENT RAILROAD GRADE CROSSING: LEVERAGING SEMANTIC SEGMENTATION AND OBJECT DETECTION FOR ENHANCED SAFETY. | | |
|---|---|---|---|
| TRACK | | | |
| Al AMIN | GRADUATE RESEARCH ASSISTANT, ELECTRICAL AND COMPUTER ENGINEERING | TENNESSEE STATE UNIVERSITY | UNITED STATES |
| Deo CHIMBA | PROFESSOR, CIVIL AND ARHITECTURAL ENGINEERING | TENNESSEE STATE UNIVERSITY | UNITED STATES |
| Kamrul HASAN | ASSISTANT PROFESSORA ELECTRICAL AND COMPUTER ENGINEERING | TENNESSEE STATE UNIVERSITY | UNITED STATES |
| Emmanuel SAMSON | GRADUATE RESEARCH ASSISTANT, ELECTRICAL AND COMPUTER ENGINEERING | TENNESSEE STATE UNIVERSITY | UNITED STATES |
| E-MAIL | aamin2@tnstate.edu;dchimba@Tnstate.edu;mhasan1@tnstate.edu; esamson@tnstate.edu; | | |

**ABSTRACT:** Crashes and delays at Railroad Highway Grade Crossings (RHGC), where highways and railroads intersect, pose significant safety concerns for the U.S. Federal Railroad Administration (FRA). Despite the critical importance of addressing accidents and traffic delays at highway-railroad intersections, there is a notable dearth of research on practical solutions for managing these issues. In response to this gap in the literature, our study introduces an intelligent system that leverages machine learning and computer vision techniques to enhance safety at Railroad Highway Grade crossings (RHGC). This research proposed a Non-Maximum Suppression (NMS)-based ensemble model that integrates a variety of YOLO variants, specifically YOLOv5S, YOLOv5M, and YOLOv5L, for grade-crossing object detection, and utilizes segmentation techniques from the UNet architecture for detecting approaching rail at a grade crossing. Both methods are implemented on a Raspberry Pi. Moreover, the strategy employs high-definition cameras installed at the RHGC. This framework enables the system to monitor objects within the Region of Interest (ROI) at crossings, detect the approach of trains, and clear the crossing area before a train arrives. In terms of accuracy, precision, recall, and Intersection over Union (IoU), the proposed state-of-the-art NMS-based object detection ensemble model achieved 96% precision. In addition, the UNet segmentation model obtained a 98% IoU value. This automated railroad grade crossing system powered by artificial intelligence represents a promising solution for enhancing safety at highway-railroad intersections.

# Intelligent Railroad Grade Crossing: Leveraging Semantic Segmentation and Object Detection for Enhanced Safety

Al Amin[1,2], Deo Chimba[1], Kamrul Hasan[1], Emmanuel Samson[1]

[1]Tennessee State University, Nashville, Tennessee, USA
[2]Email for correspondence, aamin2@tnstate.edu

1. INTRODUCTION

Railroad Highway Grade Crossing (RHGC) safety is a global concern, especially in the United States (Soleimani et al., 2019). There are approximately 212,000-grade crossings in the U.S. In 2015, there were 129,582 public crossings and 80,073 private crossings (FRA, 2015). The Federal Railroad Administration (FRA) engages in extensive engineering, education, and enforcement efforts to increase highway-rail grade crossing safety by decreasing the number, frequency, and severity of crashes each year (FRA, 2015). The US had 48,083 highway-rail grade crossing crashes between 2000 and 2018. 6,103 deaths and 18,851 injuries resulted in $302,065,336 in automobile damages (Soleimani et al., 2019). In 2019, there were 2,225 incidents at one of the more than 206,000 at-RHGCs in the United States, resulting in 297 fatalities and 815 injuries (Report, 2021). Moreover, another statistics depicts that between 2016 and 2020, an average of 2,000 crashes and 257 fatalities occurred annually at the over 200,000 highway-rail grade crossings in the United States (J. Baillargeon & Doran, 2021). Approximately 95% of all rail-related fatalities over the past decade have resulted from grade crossing collisions or right-of-way trespassing. In 2020 alone, there were 1,901-grade crossing incidents, leading to 197 fatalities and 688 injuries, in addition to 525 fatalities and 557 injuries at non-crossing locations (Bedini Jacobini & Ngamdung, 2022). Table 1 presents a comprehensive statistical report, as per the FRA, detailing the various objects, such as buses, tracks, people, etc., associated with RHGC. The report encapsulates data from the decade spanning 2010 to 2020(Wu et al., 2022).

**Table 1**. Statistical report for RHGC objects crashes between (2011-2010)(Wu et al., 2022)

| Type of Vehicle | Total Reported Crashes | Fatal | Injury | Property Damage Only (PDO) |
|---|---|---|---|---|
| **Auto** | 9477 | 6331 | 2524 | 622 |
| **Truck** | 1392 | 916 | 382 | 94 |
| **Truck-trailer** | 3450 | 2831 | 516 | 103 |
| **Pick-up truck** | 2631 | 1643 | 727 | 261 |
| **Van** | 562 | 344 | 164 | 54 |
| **Bus** | 34 | 33 | 1 | 0 |
| **School bus** | 8 | 6 | 1 | 1 |
| **Motorcycle** | 67 | 23 | 26 | 18 |
| **Other vehicles** | 1210 | 729 | 360 | 121 |
| **Pedestrian** | 30 | 2 | 7 | 21 |
| **Other** | 577 | 261 | 161 | 155 |
| **Total** | 1943813 | 119 | 4869 | 1450 |

Numerous academic and industrial research organizations across the globe are investigating several cutting-edge approaches to improve RHGC safety as well as safe traffic management in response to these challenges. These methodologies include the application of artificial intelligence (AI) and machine learning (ML) techniques(Kocbek & Gabrys, 2019)(Gibert et al., 2017)(Guo et al., 2022)(C. W. Li et al., 2014)(Dent & Marinov, 2019)(Soleimani et al., 2019)(Yoda et al., 2006)(Dagvasumberel et al., 2021)(Xiao & Liu, 2012)(Mammeri et al., 2021). The primary objective of our research is to improve safety at RHGC through the development and application of cutting-edge computer vision techniques specifically for object detection and segmentation utilizing our custom dataset. This study makes several substantial contributions to the discipline. Firstly, this research introduces a comprehensive computer vision technique for RHGC safety. For the RHGC object detection, we built an NMS (Non-Maximum Suppression) base ensemble model, which is more secure than a single algorithm. The proposed approach incorporates different object detection algorithms such as YOLOv5S, YOLOv5M, and YOLOv5L. Additionally, the proposed semantic segmentation UNet Model detects oncoming trains. These solutions enable AI-driven RHGC safety. Secondly, we have gathered real-time video data, which includes cars, buses, trucks, people, train and so on, from RHGCs and preprocesses it for incorporation into the robust object detection and segmentation models. Finally, this preprocesses dataset, which offers valuable insights into RHGC object detection research, has been made publicly available to stimulate further research in RHGC safety.

2. LITERATURE REVIEW

**Related work on obstacle detection at the RHGC Safety:**

In the United States, RHGC is a major concern for reducing crashes involving highway vehicles and people. In recent decades, global efforts from academia, industry researchers, and regulatory authorities like the FRA have contributed to improvements in the Intelligent Railroad Transportation System (IRTS) utilizing ML and DL methods (Kocbek & Gabrys, 2019)(Dent & Marinov, 2019)(Xiao & Liu, 2012)(Gibert et al., 2017). Most of this study focuses on computer vision-based image and video detection and segmentation(Dagvasumberel et al., 2021). In 2022, the FRA researcher focused on object detection and segmentation approaches for RHGC safety. They applied Mask R-CNN architecture with a ResNet50 backbone and trained on the COCO dataset for the detection of grade-crossing objects (people, automobiles, buses, trains, motorcycles, and bicycles). Both approaches obtain satisfactory Intersection over Union (IoU) performance (92% for object detection and 96% for segmentation). This research has some limitations, including a 30% false-positive rate due to insufficient segmentation overlap detection and high computational demands for video processing (Bedini Jacobini & Ngamdung, 2022). In 2021, the FRA's Automated Track Inspection Programme (ATIP) employs machine vision technology and various measurement systems to assess track conditions. It uses algorithms to detect potential defects in railroad infrastructure. ENSCO, Inc., developed software utilities for the automatic detection and extraction of high-resolution images of areas of interest. The initial results are promising for the development of an automated evaluation of switch components. However, the system requires more training data to improve detection accuracy and reduce false-positive rate (J. P. Baillargeon & Stuart, 2021). In 2021, another FRA study depicts that the Automated Video Inspection System for Grade Crossing Safety uses data ingestion, chunking, integration with an AI platform, and object detection models based on RetinaNet to find features of a railway crossing. Despite current limitations like poor video feed quality and suboptimal angles, the system's accuracy is expected to improve with better data sources. Future enhancements may incorporate a multi-camera system for enhanced feature detection(Hendricks, 2021). In the 2020 FRA, the study talks about the Automated, Drone-Based, Grade Crossing Inspection System (AXIS), which uses drones and a Convolutional Neural Network (CNN) to find features of railroad crossings with a 95% accuracy rate. A specialized flight planning app will be created, risk profile calculations will be improved, AI will be integrated for sign detection, and the system will be optimized for network-wide inspections, among other improvements(Mmary, 2020). Our research depicts a cutting-edge, NMS-based ensemble model for object detection in RHGC, integrating YOLOv5S, YOLOv5M, and YOLOv5L. This ensemble approach, surpassing previous research, obtained an impressive 96% accuracy, demonstrating its superiority over single-algorithm models.

**Related work on the segmentation approach at the RHGC Safety:**

Plenty of research is conducted at the RHGC, with a significant focus on leveraging computer vision and various segmentation models for analysis and improvement. The study examined vehicle driver injury severity factors at RHGC utilizing three models, such as the traditionally ordered logit (OL), the two-segment latent segmentation-based ordered logit (LSOL II), and the three-segment latent segmentation-based ordered logit (LSOL III). The LSOL II model obtained the most accurate fit, identifying key characteristics of "low risk" crossings. However, this research did not mention environmental factors (Eluru et al., 2012). Another research "A Y-Net deep learning method for road segmentation using high-resolution visible remote sensing images" introduces a Y-shaped convolution neural network for road segmentation. The Y-Net segments multi-scale highways while minimizing background disturbance. It achieved a mean accuracy of 79% and 82% on two testing datasets and 5-fold cross-validation, respectively. Due to the small number of road pixels and the Y-Net training period, the study experienced a class imbalance. Transfer learning may reduce computational time in the future (Y. Li et al., 2019). Another academic group study on "Cluster-Based Approach to Analysing Crash Injury Severity at RHGC" employs a two-step segmentation model to better understand crash injury severity factors at RHGC. A clustering technique and a mixed logit model obtained 70.8% and 72.9% accuracy, respectively. However, limitations include the possibility of omitted variable bias and the inability to generalize results due to location-specific characteristics (Kang & Khattak, 2017). Our study developed a UNet segmentation model, utilizing EfficientNet as the backbone, for the detection of upcoming rail. This model achieved a high validation precision of 98%.

3. METHODOLOGY

This section describes an AI-driven RHGC safety amalgamate object detection, NMS-based ensemble mode, and the segmentation model for UNet architecture for proactive train detection prior to the crossing's approach.

**Object detection and segmentation-based framework for RHGC safety and system description:**

Our proposed AI-driven framework utilizes computer vision for traffic monitoring at RHGC. The system employs Raspberry Pi interfacing with surveillance cameras for object monitoring at grade crossings. Two cameras, positioned where existing low-voltage track circuits are installed, use UNet segmentation techniques to provide real-time information about approaching trains. Four additional cameras, using an ensemble model of various YOLO variants, perform object detection tasks in the Region of Interest (ROI) at the grade crossing. The system's adaptability allows for high-accuracy traffic monitoring, with initial results indicating a processing speed of 10 frames per second. The networks can be trained to detect objects at varying highway grade crossings, enhancing their accuracy and performance. Figure 1 shows the computer vision-based framework for RHGC safety.

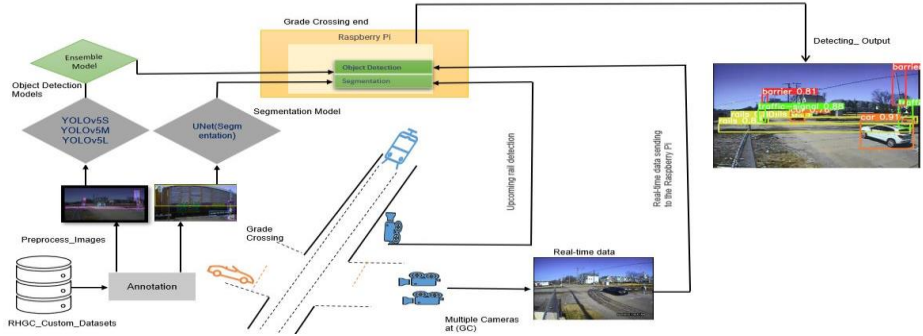
Figure 1: Computer vision based automated RHGC safety framework.

**Enhancing Object Detection in RHGC with NMS based Ensemble Method description:**

Ensemble methods are crucial in plenty of object detection research (Okran et al., 2022), particularly in grade-crossing object detection, as they amalgamate decisions from multiple models to enhance overall performance. In our research, YOLOV5S, YOLOv5M, and YOLOv5L are combined to create the NMS-ensemble model. This approach is particularly beneficial when dealing with suboptimal training data or complex hyperparameters. An NMS-based ensemble model is important for detecting people, cars, trucks, and trains on railroads and highways because it filters out overlapping bounding boxes, which improves the accuracy of detection. The NMS method considers boxes to belong to a single object if their Intersection over Union (IoU) exceeds a certain threshold. This threshold selection is critical to the model's performance. The NMS-based ensemble works by discarding redundant boxes and producing averaged localization predictions from different models. The weight update process in an NMS-based ensemble model can be mathematically represented as:

$$Wnew = W_{old} + \alpha * (y - prediction) \quad (1)$$

Where $W_{new}$ and $W_{old}$ are the new and old weights, $\alpha$ is the learning rate, y is the actual output, and prediction is the model's predicted output. Additionally, the loss function for an NMS-based ensemble model can be represented as:

$$L = \sum(y_i - \int(x_i))2 \quad (2)$$

Where L is the loss function, $y_i$ is the actual output for the i-th instance, and $\int(x_i)$ is the model's predicted output for the i-th instance. However, it's important to note that while ensemble can significantly improve performance, it may also increase the computational cost and inference time.

**NMS based Ensemble model IoU calculation procedure:**

In this research, an NMS-based ensemble model calculates the IoU value by dividing the area of overlap between the predicted and ground truth bounding boxes by the area of union. IoU is the traditional statistical evaluation criterion for object detection studies. It evaluates how closely the anticipated bounding box matches the ground truth. In the below picture, we have a green box (r) and a grey box (b). The bounding box r has $(r.x_1, r.y_1)$ and $(r.x_2, r.y_2)$. Another bounding box b has $(b.x_1, b.y_1)$ and $(b.x_2, b.y_2)$.

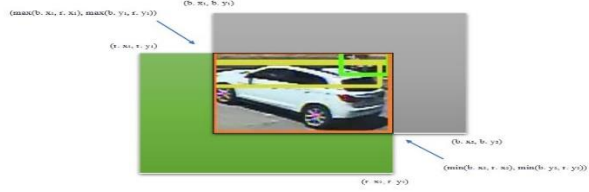

Figure 2: IoU calculation for RHGC object.

To calculate the intersection (object) area, two coordinates where the edges of two boxes intersect must be known:

$$Width = \min(b.X_2, r.X_2) - \max(b.X_1, r.X_1) \quad (3)$$
$$Height = \min(b.y_2, r.y_2) - \max(b.y_1, r.y_2) \quad (4)$$

Therefore, the intersection is the multiplication of Equations 3 and 4.
$Intersection = width \times Height$
The union area of two boxes is:
Union= $r_{area} + b_{area} - intersection$. We must deduct the intersection. Otherwise, the area of the object (car) would be duplicated. Finally, the IoU can be computed as follows:

$$IoU = \frac{Intersection}{Union} \quad (5)$$

**Why our research incorporates the YOLO5 variant to build an ensemble model for the final prediction of RHGC safety:**

The integration of YOLOv5 model variants (YOLOv5S, YOLOv5M, and YOLOv5L) into the RHGC safety research presents a formidable ensemble model final solution for real-time object detection, a critical component in enhancing the safety and efficiency of these crossings in Figure 1. YOLO5, a cutting-edge object identification system, can quickly identify, classify, and locate objects in photos and videos. Our real-time grade-crossing object identification research uses a custom video dataset. Real-time object detection in RHGC, where vehicles, pedestrians, and other things must be detected quickly and accurately, is appropriate for its 140 frames per second processing. YOLOv5's high mean average precision (mAP) ensures excellent item detection and classification, which is crucial for identifying and controlling crossing dangers. YOLOv5 squares an image or video frame without changing the aspect ratio. It detects low-level features like edges and colours and high-level features like forms and objects using convolutional layers. The system predicts object locations using anchors and bounding boxes, with k-means clustering determining anchor sizes. The head of the model, consisting of three output layers, each responsible for detecting objects of different sizes, outputs the predictions. The backbone of the model, responsible for feature extraction, is a modified version of the CSPDarknet53, a convolutional neural network designed for efficient feature extraction. The proposed ensemble model's infrastructure, shown in Figure 3, shows how individual base models are trained using a training dataset. We constructed the NMS-Ensemble model for final prediction using the testing dataset by combining those base model weights.

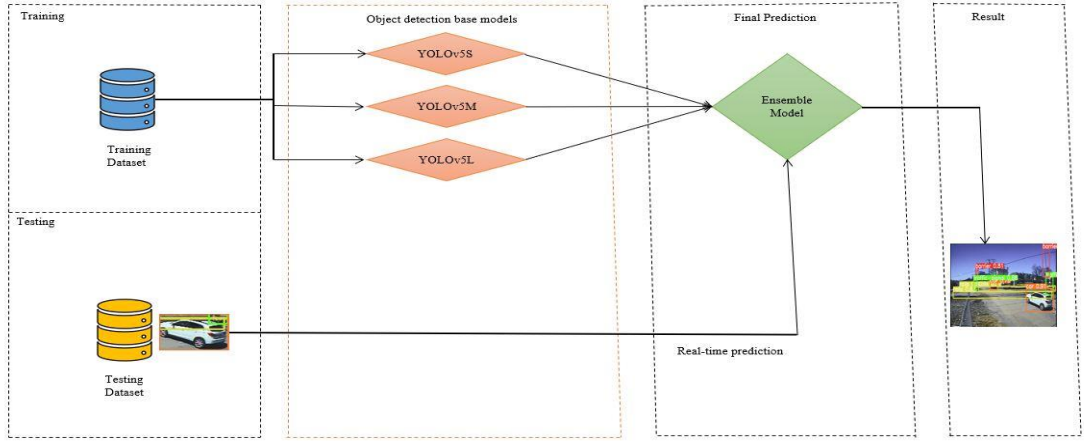

Figure 3: The Architecture of the Proposed Ensemble Model Integrating YOLOv5 Variants for Real-Time Object Detection at RHGCs.

**Architectural description of segmentation techniques (UNet) for detecting approaching rails for RHGC safety:**

The U-Net model is renowned for its segmentation capabilities in different industries, such as healthcare research, industrial defect detection, and automobile industries, which are employed in our research for detecting upcoming rail semantic segments for RHGC safety (Joo et al., 2022)(Amin et al., 2023)(Do et al., 2021). The model operates on 224x224 RGB inputs and utilizes EfficientNet as its encoder component. This encoder processes the input, creating feature maps of varying scales that reduce spatial dimensions while increasing depth. Following this, components forming the U-Net decoders, Sequential, Concatenate, Sequential_1 to Sequential_3, and Concatenate_1 to Concatenate_3, progressively enlarge the spatial dimension while reducing the depth. This process of extraction and reconstruction, facilitated by skip connections, helps preserve high-resolution features, which in turn improves segmentation results. The last layer, conv2d_transpose_4 (Conv2DTranspose), resizes the feature map to match the input image size, generating the segmentation map. Figure 4 shows the UNet model layer representation for RHGC safety.

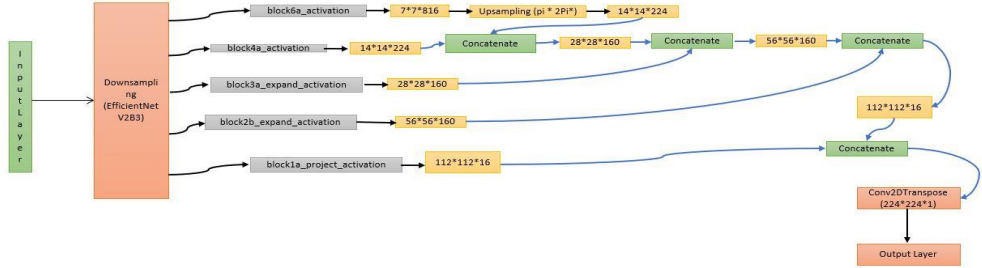

Figure 4: UNet Segmentation model Layer representation at RHGC safety

**Semantic segmentation loss function (Binary cross entropy) at RHGC safety:**

In our research on the safety of RHGC, we employ binary cross entropy (BCE) as the loss function for our semantic segmentation task. The BCE loss function computes the cross-entropy loss between the true labels and predicted labels, making it appropriate for binary classification problems. It quantifies the difference between the predicted probability distribution and the actual distribution, allowing the model to learn to segment the impending rail segments in images accurately. By minimizing this loss during training, we improve the model's ability to detect approaching railroads, thereby enhancing RHGC's safety. Cross-entropy is the difference between two probability distributions for a random variable or set of events. As segmentation is a form of classification at the pixel level, it is extensively used for classification purposes and is effective. Binary Cross-Entropy is defined as:

$$L_{BCE}(y, \hat{y}) = -(y \log(\hat{y}) + (1 - y) \log(1 - \hat{y})) \qquad (6)$$

Here, $\hat{y}$ is the predicted value by the prediction model.

## 4. AI-DRIVEN COMPUTER VISION-BASED SOLUTION APPROACHES FOR RHGC SAFETY

Our research presents a two-pronged computer vision strategy for enhancing safety at RHGCs. It employs a three-YOLO variant (YOLOv5S, YOLOv5M, and YOLOv5L) ensemble model for object detection and a UNet model for semantic segmentation. Real-time video feeds from RHGC cameras are processed to detect approaching trains and monitor the crossing area for obstructions. Typically, the object detection model evaluates a region of ROI that contains the railway crossing. If no obstructions are detected, the system lowers the level crossing bar; if obstructions are detected, a changeable message sign (CMS) displays a warning message. This dual approach provides comprehensive surveillance, which substantially contributes to the improvement of RHGC safety. Algorithm 1 detecting objects at grade crossing by utilizing tower mounted cameras and ensemble model. Algorithm 2 process presence of upcoming rail at grade crossing by utilizing segmentation technique (UNet) model.

**Algorithm 1:** Algorithmic Procedure of AI-driven (Object Detection) an automated RHGC safety

**Data:** RGC_dataset.mp4, grade_crossing_locations
**Result:** Decision on level bar control and warning message broadcast

```
1  Function Main(RGC_dataset.mp4, grade_crossing_locations):
2      dataset ← cv2.VideoCapture('RGC_dataset.mp4')
3      Annotate Images: This step is typically done utilizing an annotation
       tool robot flow.
4      model1 ← YOLO.train(dataset, version=5S)
5      model2 ← YOLO.train(dataset, version=5M)
6      model3 ← YOLO.train(dataset, version=5L)
7      ensemble_model ← Ensemble Model(model1, model2, model3)
8      Implemented ← Raspberry Pi()
9      Raspberry Pi.train_model(ensemble_model)
10     Raspberry Pi ← [Server(Raspberry Pi, location) for location in
          grade_crossing_locations]
11     Collect Real-Time Video: This step involves the real-time collection
       of video data from cameras installed at the grade crossings.
12     for Raspberry Pi_servers do
13         video_feed ← server.get_video_feed()
14         objects_detected ← ensemble_model.predict(video_feed)
15         if server.is_clear(objects_detected) then
16             server.lower_level_bar()
17         else
18             server.raise_level_bar()
19         if server.is_not_clear(objects_detected) then
20             CMS.broadcast_warning_messages(server.get_location())
21         end
22     end
23     Grade Crossing Safety:
```

**Algorithm 2:** Algorithmic Procedure for AI-Driven(Segmentation Techniques) for Automated Railroad Grade Crossing Safety

**Data:** camera1, camera2, threshold, ROI
**Result:** Decision on level bar control and warning message broadcast

```
1  Function Main(camera1, camera2, threshold, ROI):
2      camera1_feed ← get_camera_feed(camera1)
3      camera2_feed ← get_camera_feed(camera2)
4      unet_model ← load_unet_model()
5      train_detected_camera1 ← unet_model.predict(camera1_feed) > threshold
6      train_detected_camera2 ← unet_model.predict(camera2_feed) > threshold
7      if train_detected_camera1 or train_detected_camera2 then
8          object_detection_model ← load_object_detection_model()
9          ROI ← get_ROI()
10         objects_detected ← object_detection_model.predict(ROI)
11         if is_clear(objects_detected) then
12             lower_level_crossing_bar()
13         else
14             CMS.broadcast_warning()
15         end
16     else
17         No train detected. Continue normal operation.
18     end
```

## 5. DATA COLLECTION FOR RHGC SAFETY

We used the Miovision Scout®, a rugged, weather-resistant camera-based traffic data collection equipment, to capture real-time video data from numerous RHGCs in Nashville, Tennessee, USA, including 300 Williams Avenue, Madison, and 210 Nesbitt Lane, Madison. Vehicles, buses, trains, people, and trucks were detected at 10-hour intervals on different days. OpenCV converted video input into frames, and Roboflow, a powerful image annotation tool, annotated each object of interest. Roboflow's transformations—rotation, flipping, mosaic, crop, and scaling—enhanced our model's robustness. YOLO data was exported for algorithm training. This thorough method assured data quality and consistency, which impacted our model's performance.

## 6. EXPERIMENTAL RESULT ANALYSIS FOR AI- DRIVEN MODELS AT THE RHGC SAFETY

**Ensemble model performance analysis for RHGC safety:**

Our research evaluates the NMS ensemble model's safety performance in RHGC using several machine-learning metrics. The confusion matrix, precision-recall curve, training and validation loss, and precision are examples. As illustrated in Figure 5, the model's confusion matrix shows impressive true positive rates of 95% and 94% for the train and people classes, respectively. The train has a 0.06% false-positive rate and a 0.02% false-negative rate. Also, the person has a 0% false-positive rate and a 0.06% false-negative rate. These findings illustrate the model's accuracy in object identification and classification; its overall precision is 96%. In class imbalance cases, the precision-recall curve is critical in figure 6. By showing the precision-recall trade-off, this curve helps explain the model's performance. Equation 7 and Equation 8 represent the precision and recall of mathematical expressions, and by combining both equations, we calculated the F1 score for each class in Equation 8. As an example of identifying car class F1 score for the RHGC.

$$Precision = (Class = car): \frac{TP(Class = car)}{TP(Class = car) + FP(Class = car)} \quad (7)$$

$$= \frac{92}{92 + 0.57} = 99\%$$

$$Recall(Class = Car): \frac{TP(Class = car)}{TP(Class = car) + FN(Class = car)} \quad (8)$$

$$= \frac{92}{92 + 0.03 + 0.05} = 99\%$$

In terms of detecting car, to find $F_1$ score, we apply this formula:

$$F_1\ score(Class = car) := \frac{2 \times Precision\ (Class = car) \times Recall(Class = Car)}{Precision\ (Class = car) + Recall(Class = Car)} \quad (9)$$

$$= \frac{2 \times 0.99 \times 0.99}{0.99 + 0.99} = 99\%$$

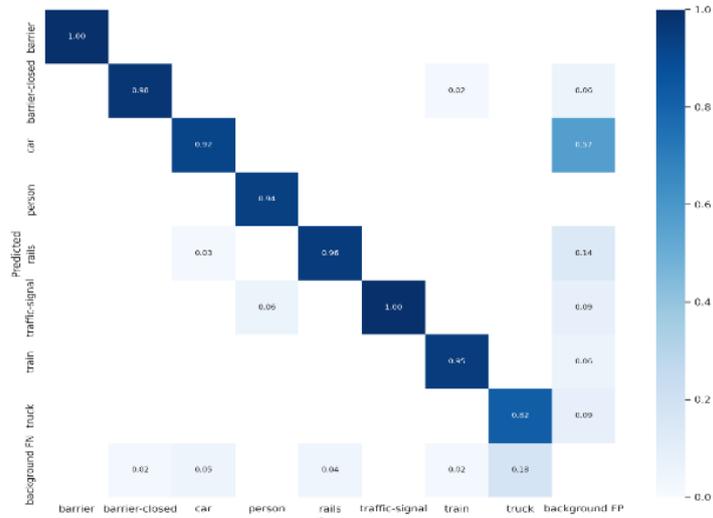

Figure 5: Ensemble model confusion matrix for detecting object at RHGC.

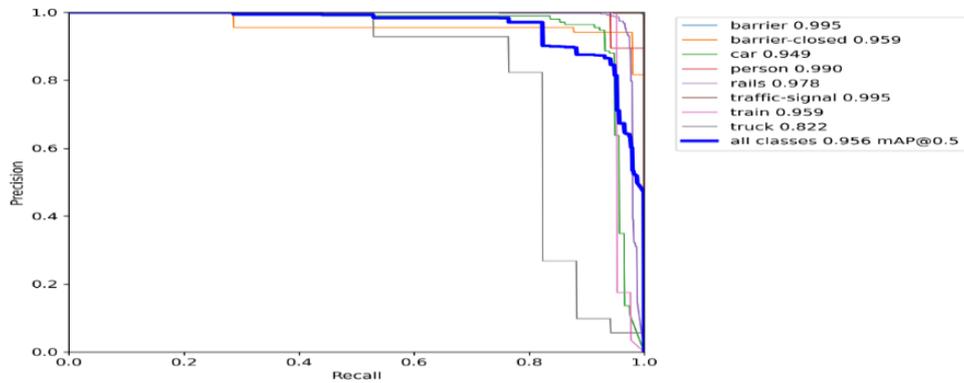

Figure 6: Ensemble model precision- recall curve at the RHGC object detection

Figures 7 and 8 show our model's performance indicators for object detection and UNet segmentation respectively: training and validation loss and accuracy curves. These curves help identify overfitting, optimise hyperparameters, track training progress, and compare models. Time-decreasing losses suggest optimal learning. Extensive RHGC safety testing validates our model. These findings advance this field and provide a solid platform for future research.

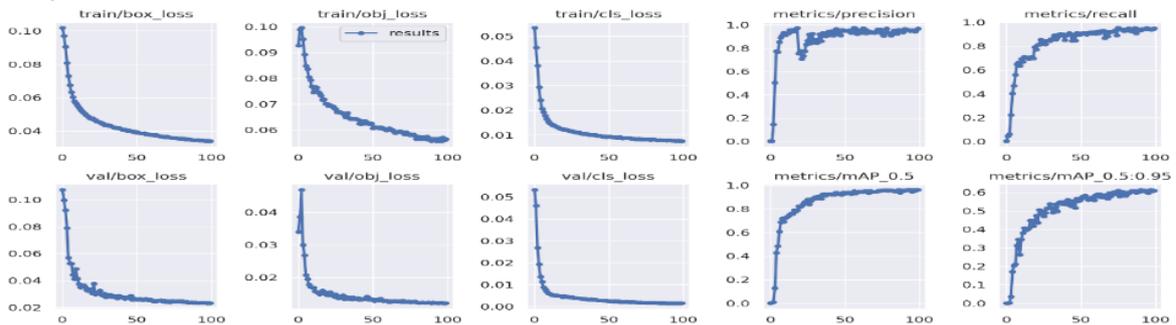

Figure 7: Ensemble model training and validation loss at the RHGC object detection.

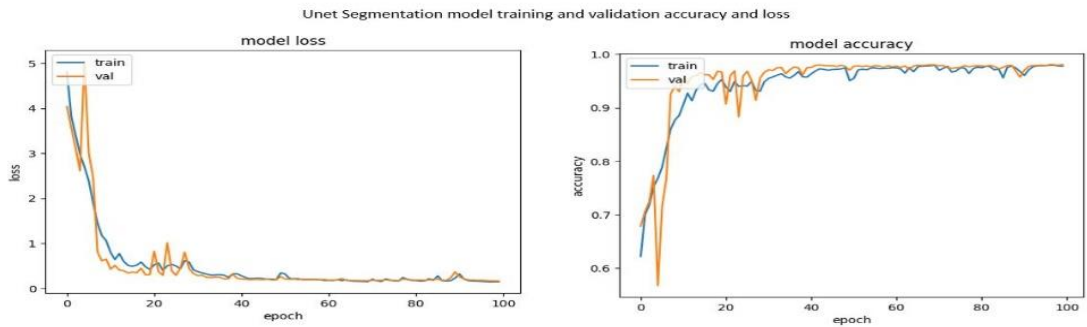

Figure 8: UNet Segmentation model for approaching rail detecting at RHGC.

**Analysis of segmentation model performance for RHGC safety:**

The UNet segmentation model is employed for segmenting test images. Pixels with values exceeding 0.5 are classified as the object (train), while those with values of 0.5 or less are deemed the background. The Intersection over Union (IoU), a metric quantifying the overlap between the predicted and ground-truth segmentation, is used to assess model performance. The Mean IoU, computed across all classes using TensorFlow's function, is approximately 98%, indicating a high degree of overlap and thus, superior model performance. Figure 9 provides a grayscale representation of the predicted segmentation, the original image, and the ground truth, with white and black areas denoting the rail and background, respectively. This visual evidence further substantiates the model's exceptional performance.

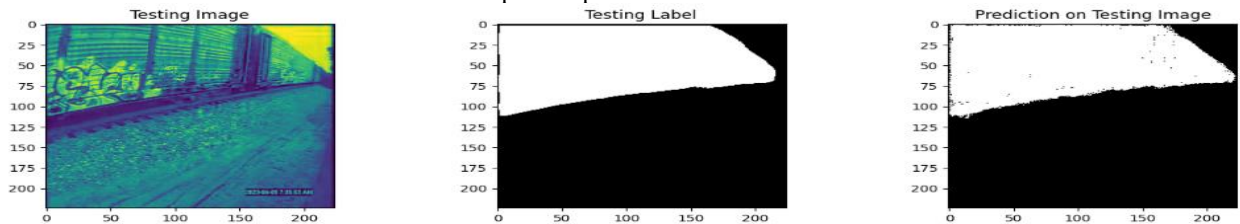

Figure 9: Comparative Analysis of Predicted Segmentation, Original Image, and Ground Truth

7. CONCLUSION AND FUTURE WORK

This research introduces a robust AI solution employing object detection and segmentation to enhance safety at RHGCs. The proposed computer vision-based framework utilizes an ensemble technique for object detection and a UNet model for rail detection, achieving 96% and 98% precision, respectively. It sets a new benchmark in RHGC safety and efficiency. Future work should consider incorporating data under varied weather conditions for broader applicability and improving the CMS for better grade-crossing communication. Exploring other ensemble techniques and alternative algorithms can further enhance model accuracy.

8. ACKNOWLEDGEMENTS

This work was generously supported by the FRA and facilitated by the facilities at the Tennessee State University Engineering Lab. We aim to continue contributing to the critical area of RHGC safety.